\documentclass{article}
\usepackage[latin9]{inputenc}
\usepackage{color}
\usepackage{mathtools}
\usepackage{amsmath}
\usepackage{amsthm}
\usepackage{amssymb}
\usepackage{graphicx}
\usepackage{microtype}
\usepackage{svg}
\usepackage[unicode=true,
 bookmarks=false,
 breaklinks=false,pdfborder={0 0 1},backref=section,colorlinks=false]
 {hyperref}

\makeatletter

\providecommand{\tabularnewline}{\\}

\usepackage{subfigure}
\usepackage{booktabs}
\usepackage{enumitem}
\setlist[itemize]{topsep=0pt,partopsep=0pt,noitemsep,leftmargin=*}
\setlist[enumerate]{topsep=0pt,partopsep=0pt,noitemsep,leftmargin=*}

\usepackage[accepted]{icml2024}

\usepackage[capitalize,noabbrev]{cleveref}

\theoremstyle{plain}

\theoremstyle{definition}
\theoremstyle{remark}
\theoremstyle{plain}
\newtheorem{thm}{\protect\theoremname}\theoremstyle{definition}
\newtheorem{problem}[thm]{\protect\problemname}\theoremstyle{remark}
\newtheorem{rem}[thm]{\protect\remarkname}\theoremstyle{remark}
\newtheorem{claim}[thm]{\protect\claimname}

\providecommand{\claimname}{Claim}
\providecommand{\problemname}{Problem}
\providecommand{\remarkname}{Remark}
\providecommand{\theoremname}{Theorem}
\usepackage[textsize=tiny]{todonotes}

\icmltitlerunning{Arrows of Time for Large Language Models}

\usepackage{caption}
\makeatother

\providecommand{\claimname}{Claim}
\providecommand{\problemname}{Problem}
\providecommand{\remarkname}{Remark}
\providecommand{\theoremname}{Theorem}

\begin{document}
\twocolumn[
\icmltitle{Arrows of Time for Large Language Models}

\icmlsetsymbol{equal}{*}

\begin{icmlauthorlist}
\icmlauthor{Vassilis Papadopoulos}{equal,EPFL,FSL}
\icmlauthor{J\'er\'emie Wenger}{gold}
\icmlauthor{Cl\'ement Hongler}{equal,EPFL}
\icmlcorrespondingauthor{Cl\'ement Hongler}{clement.hongler@epfl.ch}
\end{icmlauthorlist}

\icmlaffiliation{EPFL}{CSFT/Institute of Mathematics, EPFL, Lausanne, Switzerland}
\icmlaffiliation{gold}{Department of Computing, Goldsmiths/University of London, London, UK}
\icmlaffiliation{FSL}{FSL/Institute of Physics, EPFL}
\icmlkeywords{Machine Learning, ICML}

\vskip 0.3in
]



\printAffiliationsAndNotice{\icmlEqualContribution} 

\begin{abstract}
We study the probabilistic modeling performed by Autoregressive Large Language
Models (LLMs) through the angle of time directionality, addressing a question first raised in \cite{shannon_prediction_1951}. For large enough models, we empirically find a time asymmetry in their ability to learn natural
language: a difference in the average log-perplexity when trying to predict the next token versus when trying to predict the previous one.
This difference is at the same time subtle and very consistent across various modalities (language, model size,
training time, ...).
Theoretically, this is surprising: from an information-theoretic point of view, there should be no such difference.
We provide a theoretical framework to explain how such an asymmetry can appear from sparsity and computational complexity considerations, and outline a number of perspectives opened by our results.
\end{abstract}
\section{\label{sec:introduction}Introduction}

Generative Models have revolutionized modern AI, yielding a wide array of
applications. Modern works have shown that such models can perform
spectacularly (and somewhat mysteriously) well on various kinds of data. Text
is perhaps the domain where progress has been the most drastic: in a few years,
Large Language Models (LLMs) have gone from generating barely correct sentences
to producing consistent stories, code, and performing countless new tasks; key
milestones include the Transformer architecture \cite{vaswani_attention_2017},
BERT \cite{devlin_bert_2019}, and GPTs \cite{radford_improving_2018,
radford_language_2019, brown_language_2020, openai_gpt-4_2023}.

At the heart of these developments are probabilistic models trained in an
unsupervised manner on vast amounts of data, for prediction or recovery tasks:
this yields an estimation of the probability measure underlying the data. These
probabilistic models appear to gain surprising abilities, such as reasoning, as
their sizes increase (see \cite{wei_emergent_2022, schaeffer_are_2023} among
others).

In this work, we investigate the interplay between the probabilistic structure
of autoregressive LLMs and the data they are trained on. More precisely, we
investigate how time directionality influences their
ability to model natural and synthetic languages.

\subsection{\label{subsec:autoregressive-llms}Autoregressive LLMs}

Famously, the pre-training of LLMs such as the GPTs consists in `learning to
predict the next token' knowing previous ones, in sequences extracted from
large text corpora, using the natural time ordering of the data they are
being trained on. A vocabulary $\mathcal{V}$ of $V$ tokens is chosen; the dataset is
then tokenized into a sequence of tokens in $\mathcal{V}$; at each step, the
model reads a sequence of tokens and outputs a probability distribution on
$\mathcal{V}$ predicting the next token.

Typically, as a probabilistic model, an autoregressive model will
estimate the probability that $n$ random consecutive tokens
$\left(X_{1},\cdots,X_{n}\right)$ are equal to
$\left(x_{1},\cdots,x_{n}\right)\in\mathcal{V}^{n}$ by taking the product of
the (estimations of) the probabilities
\begin{align}
    \mathbb{P} & \left\{X_{1}=x_{1}\right\} \nonumber \\
    \mathbb{P} & \left\{X_{2}=x_{2}|X_{1}=x_{1}\right\} \nonumber \\
    \vdots\label{eq:fw-prob-model} \\
    \mathbb{P} & \left\{X_{n}=x_{n}|X_{1}=x_{1},\cdots,X_{n-1}=x_{n-1}\right\} ,\nonumber
\end{align}
yielding an estimated probability measure
$\mathbb{P}_{n}^{\rightarrow}$ on $\mathcal{V}^{n}$.

Autoregressive LLMs (GPTs, GRUs, LSTMs, \dots) thus factorize (their estimates
of) the joint probabilities in terms of conditional probabilities for each
token knowing past ones. This brings a number of advantages: first, this
leverages the fact that for each token sequence $x_{1},\ldots,x_{n}$, each
token $x_{k}$ is used in to predict each token $x_{\ell}$ with $\ell>k$. In
particular, GPT (compared to the earlier BERT) includes causality-aware
attention, allowing for a parallelization of the training process: a sequence
$x_{1},\ldots,x_{n}$ generates $n-1$ fully parallelizable tasks (predict
$x_{k}$ from $\left(x_{1},\ldots,x_{k-1}\right)$ for $2\le k\leq n$). Also,
this representation enables a natural sampling from
$\mathbb{P}_{n}^{\rightarrow}$ (token by token), as well as data compression:
the factorization decomposes these processes into many smaller substeps
\cite{graves_bayesian_2023}.

Autoregressive LLMs such as GPTs have enabled a massive scaling up of the
number of parameters and dataset sizes, yielding numerous fascinating
phenomena, e.g. scaling laws \cite{kaplan_scaling_2020, hoffmann_training_2022}
and emergent behavior, e.g. abilities at arithmetic operations \cite{shen_positional_2023}, circuit
computing tasks \cite{dascoli_boolformer_2023}, or high-level linguistic
proficiency.

\subsection{\label{subsec:arrow-of-time-problem}Arrow of Time and Language Models}

While decomposing measures into a sequence of conditional probabilities is
natural, it is not a priori clear why following the time direction of language
to do so is optimal (except for downstream tasks, e.g. making a chatbot): what
is the best order when predicting the token probabilities? A natural idea to
investigate this question is simply to reverse the Arrow of Time: to
estimate probabilities \emph{backward}. This amounts to training models on
time-flipped datasets: we train the same models on the same data slices for
next-token predictions, but for each data slice
$\left(x_{1},\cdots,x_{n}\right)$ we feed the model with
$\left(x_{n},x_{n-1},\cdots,x_{1}\right)$ instead.

As a result, instead of (\ref{eq:fw-prob-model}), we take the
product of the estimations of
\begin{align}
    \mathbb{P} & \left\{ X_{n}=x_{n}\right\} \nonumber \\
    \mathbb{P} & \left\{ X_{n-1}=x_{n-1}|X_{n}=x_{n}\right\} \nonumber \\
    \vdots\label{eq:bw-prob-model}\\
    \mathbb{P} & \left\{ X_{1}=x_{1}|X_{n}=x_{n},\cdots,X_{2}=x_{2}\right\} .\nonumber
\end{align}
This yields an estimated probability measure $\mathbb{P}_{n}^{\leftarrow}$
on $\mathcal{V}^{n}$.

In this paper, we will speak of \emph{forward/backward (FW/BW) model} to refer
to the same (architectural) model trained with the same hyperparameters
(learning rate, batch size, training time, ...) but fed with (batches of)
$\left(x_{1},\ldots,x_{n}\right)$ and $\left(x_{n},\ldots,x_{1}\right)$ from
the same dataset respectively. In other words, both models are the same, except
that the FW model is trained to predict the \emph{next} token, while the BW one
is trained to predict the \emph{previous} token.

\begin{problem}
    For a measure $\mathbb{P}$ and a given model, how do the \emph{forward}
    \emph{and backward measures} $\mathbb{P}_{n}^{\rightarrow}$ and
    $\mathbb{P}_{n}^{\leftarrow}$ differ from one another?
\end{problem}

For certain $\mathbb{P}$s, we will see universal asymmetries: for
any given architecture and hyperparameters, a substantial difference
between the way $\mathbb{P}_{n}^{\rightarrow}$ and $\mathbb{P}_{n}^{\leftarrow}$
approximate $\mathbb{P}$ arises.

\subsection{\label{subsec:cross-entropy-loss-and-perplexity}Cross-Entropy Loss
and Perplexity}

LLMs are trained as follows: sample sequences of $n$ consecutive tokens
$\left(x_{1},\ldots,x_{n}\right)$ from the dataset; then, for $i=1,\ldots,n$, get a prediction $p_{i}:\mathcal{V}\to\left[0,1\right]$ for $X_{i}$ (given
previous tokens for the FW model/next tokens for the BW one), compute
the loss $\sum_{i=1}^{n}\ell\left(p_{i},x_{i}\right)$ on the observed tokens
$x_{i}$ for a loss function $\ell$, perform a gradient step to
optimize $\ell$, and start again.

In the training of most LLMs, the prime choice for $\ell$ is the
\emph{cross-entropy loss}, defined by
$\ell_{\mathcal{C}}\left(\mathbf{p}_{k},x_{k}\right)=-\ln\mathbf{p}_{k}\left(x_{k}\right)$:
the negative log of the \emph{predicted probability of the observed token}. It
is a \emph{proper scoring rule} \cite{savage_elicitation_1971,
gneiting_strictly_2007}, uniquely identified by certain modularity properties
\cite{hanson_logarithmic_2012}; in expectation, it gives the number of nats
($\ln{2}$ times the number of bits) needed to compress
$\left(x_{1},\ldots,x_{n}\right)$ when using a coding scheme based on the
model's estimated probabilities. Finally, and crucially for us, we have the
following:

\begin{rem}\label{rem:log-perp-repr}
For $i=1,\ldots,n$, let $\left(\mathbf{p}_{i}^{\rightarrow}\right)_{i}$
and $\left(\mathbf{p}_{i}^{\leftarrow}\right)_{i}$ denote the predictions
of the FW and BW models respectively. Setting $\ell_{i}^{\leftrightarrows}:=\ell_{\mathcal{C}}\left(\mathbf{p}_{i}^{\leftrightarrows},x_{i}\right)$,
we have
\[
\sum_{i=1}^{n}\ell_{i}^{\rightleftarrows}=-\ln\mathbb{P}_{n}^{\leftrightarrows}\left\{ X_{1}=x_{1},\cdots,X_{n}=x_{n}\right\} .
\]
In particular, if the FW and BW measures coincide, the cross-entropy
losses are identical.
\end{rem}

\begin{rem}
\label{rem:log-perp-and-cross-entropy}If $\left(x_{1},\ldots,x_{n}\right)$
is sampled from $\mathbb{P}_{n}$, denoting by $\mathcal{L}_{n}^{\leftrightarrows}$
the expectations of $\sum_{i=1}^{n}\ell_{i}^{\leftrightarrows}$ (estimated
by the test loss of the models during training), we have
\[
\mathcal{L}_{n}^{\leftrightarrows}=\mathrm{D}_{\mathrm{KL}}\left(\mathbb{P}_{n}\big|\big|\mathbb{P}_{n}^{\leftrightarrows}\right)+H\left(\mathbb{P}_{n}\right),
\]
where $H$ denotes the entropy and $\mathrm{D}_{\mathrm{KL}}$ the
Kullback-Leibler divergence.
\end{rem}

\begin{rem}
\label{rem:example-log-perp-comp}It is worth noting that, in spite of its apparent triviality, Remark
\ref{rem:log-perp-repr} crucially depends on the
choice $\ell$ as $\ell_{\mathcal{C}}$. Moreover, even if
$\mathbb{P}_{n}^{\rightarrow}=\mathbb{P}_{n}^{\leftarrow}$, we will generally
have $\ell_{i}^{\rightarrow}\neq\ell_{i}^{\leftarrow}$ for $1\leq i\leq n$
(though $\sum_{i}\ell_{i}^{\rightarrow}=\sum_{i}\ell_{i}^{\leftarrow}$). When
$\mathbb{P}_{n}^{\rightarrow}=\mathbb{P}_{n}^{\leftarrow}=\mathbb{P}_{n}$ we
have that $\ell_{i}^{\rightarrow}$ and $\ell_{i}^{\leftarrow}$ yield two
(typically different) decompositions of the log-likelihood of
$\left(x_{1},\ldots,x_{n}\right)$. For instance, take as a dataset the 81
expressions $A\times B=CD$ for $1\leq A,B\leq9$, $0\leq C,D \leq 9$ (setting $C=0$ when needed).
The FW log-perplexity is concentrated on $A$ and $B$, each contributing
$\ln{9}\approx2.2$ nats:
$\ell_{A}^{\rightarrow}=\ell_{B}^{\rightarrow}\approx2.2$,
$\ell_{C}^{\rightarrow}=\ell_{D}^{\rightarrow}=0$. The BW log-perplexity is
distributed differently: for instance, for $3\times4=12$,
$(\ell_{A}^{\leftarrow}, \ell_{B}^{\leftarrow}, \ell_{C}^{\leftarrow},
\ell_{D}^{\leftarrow}) \approx (0, 1.39, 1.1, 1.91)$.
\end{rem}

\begin{rem}
Remark \ref{rem:log-perp-and-cross-entropy} suggests that if
$\mathbb{P}_{n}^{\rightarrow}$ and $\mathbb{P}_{n}^{\leftarrow}$ coincide (e.g.
if both models have learned the true measure $\mathbb{P}$, memorized the
training set, or more generally have learned $\mathbb{P}$ `equally well'), their
associated average losses should be equal. If we take
a very small dataset or context length, we can expect to have
$\mathcal{L}_{n}^{\rightarrow}\approx\mathcal{L}_{n}^{\leftarrow}$. If we are
to train a FW and a BW model with our setting, any substantial difference in
their cross-entropy losses \emph{will necessarily reflect an asymmetry of
$\mathbb{P}$} (w.r.t. its learnability by the models)\emph{.}
\end{rem}

As it will turn out, for many types of data (i.e. $\mathbb{P}$s), a consistent
difference between FW and BW log-perplexities arises across a wide range of models
and hyperparameters.

\subsection{\label{subsec:setup-and-plan}Setup and Plan}

In Section \ref{subsec:cross-entropy-loss-and-perplexity}, we showed that a
difference between FW and BW losses reflects a difference between the measures
$\mathbb{P}_{n}^{\rightarrow}$ and $\mathbb{P}_{n}^{\leftarrow}$ learned by the
FW/BW models, all else being equal: same dataset, same model (architecture and
hyperparameters). In such a case, we say that $\mathbb{P}$ (or the
corresponding dataset) exhibits an \emph{Arrow of Time (AoT)} with respect to
the model and context length $n$. We speak of a \emph{FW AoT} if the average FW
log-perplexity is below the BW one (i.e. if the FW model outperforms the BW
one).

This paper aims to investigate the following questions:
\begin{itemize}
\item Is there an AoT in large natural language datasets? Does it depend on the
    language? Does it depend on the context length $n$?
\item Can we formulate a theoretical framework explaining the presence of an
    AoT? Can we construct simple synthetic datasets exhibiting an AoT? Can the
    presence of an AoT be explained mathematically from first principles? What
    should we expect as the model sizes tend to infinity?
\end{itemize}

The paper is organized as follows:

\begin{itemize}
\item In Section \ref{sec:empirical-results-on-nat-lang}, we investigate the
    existence of an AoT, starting from a basic setup and expanding across
    modalities: languages, architectures,
    hyperparameters, context  lengths.
\item In Section \ref{sec:computability-and-irreversibility}, we investigate
    the theoretical origins of AoTs, starting with a simple synthetic
    dataset exhibiting one; in this case, the difference between
    $\mathcal{L}_{n}^{\rightarrow}$ and $\mathcal{L}_{n}^{\leftarrow}$ can be
    shown to be related to the hardness of the factoring problem (Section
    \ref{subsec:number-factoring}). We then introduce a more general class of
    synthetic datasets which we call `linear languages' (Section
    \ref{subsec:binary-operations}), providing us with a fairly wide class
    of datasets with a mathematically justified AoT.
\item Finally, in Section \ref{sec:discussion}, we summarize our results and
    outline a number of possible future research directions.
\end{itemize}

\subsection{\label{subsec:relation-to-previous-work}Relation to Previous Works in Language Modeling}

To the best of our knowledge, the question of comparing FW and BW text generation in Language Modeling was first raised in
\cite{shannon_prediction_1951}: Shannon ran experiments on the task of predicting the next vs previous letters, noting the theoretical equality between FW and BW entropies; he noted that while the BW prediction appeared to be ``subjectively much more difficult'' for humans, it led to ``only slightly poorer'' scores. 

A notable recent example is \cite{sutskever_sequence_2014}, focusing on machine
translation using LSTMs, finding that reversing the source sentence (i.e. training the source LSTM backwards) improves performance. Other well-known
examples of related techniques include ULMFiT \cite{howard_universal_2018} and
ELMO \cite{peters_deep_2018}, the already mentioned BERT \cite{devlin_bert_2019},
as well as T5 \cite{raffel_exploring_2023}, and XLNET \cite{yang_xlnet_2020}.

Attempts at combining FW/BW models include \cite{mou_backward_2016,
liu_agreement_2016, zhang_regularizing_2018, serdyuk_twin_2018,
mangal_lstm_2019} (using RNNs); or recently \cite{nguyen_meet_2023}, a `Meet in
the Middle' approach which shows how pre-training using FW/BW Transformer
models enhance FW-only autoregressive generation; as well as in
\cite{shen_positional_2023}, applying the idea of reversing data to improve LLM
performance (see also Section \ref{subsec:number-factoring}). In these
approaches, the FW/BW models are treated as one model, yielding one combined
loss. They compare various models on a task (see section 5 of
\cite{nguyen_meet_2023} for a comprehensive review), rather than study
potential discrepancies in FW/BW learning.

Some results showing BW models performing equally or even better than FW models can be found in the literature. While \cite{vinyals_order_2016} highlights the importance of order
(of input and output sequences) for performance, and shows that scrambling words reduces performance, it shows FW and BW seemingly performing equally well. 
In a recent work \cite{pfau_eliciting_2023} use BW GPT models to perform adversarial attacks on LLMs, showing slightly better accuracy BW than FW. 
An older study \cite{duchateau_confidence_2002} based on trigram models, also seemingly 
reports better BW than FW performances. Note that these works do not affect our confidence in our results, given the magnitude of our experiments and the level of care involved in their setup. 

A number of works, in particularly related to the machine-translation setup, try to use token re-ordering to improve performances in one way or another, see, e.g. \cite{wu_beyond_2018, oord_parallel_2017, gu_non-autoregressive_2018, lee_deterministic_2018, ford_importance_2018, savinov_step-unrolled_2022, welleck_non-monotonic_2019, stern_insertion_2019, gu_levenshtein_2019, chan_empirical_2019, chan_kermit_2019, gu_insertion-based_2019, emelianenko_sequence_2019, mansimov_generalized_2020}. See \cite{xiao_survey_2023} for a survey. Given the extensive body of work on the topic, it comes across as somehow surprising that the effect we highlight in our paper is not noted anywhere (besides in the early works of Shannon). Among possible reasons for this could be the use of translation-specific metrics such as the BLEU score \cite{papineni_bleu_2001}, rather than cross-entropy losses, and the lack of careful setups comparing FW and BW performance for large models on large datasets, all else being equal.  
\subsection{\label{subsec:relation-to-previous-theoretical-work}Causality and Information Theory}
While the AoT effect we highlight in this paper is surprising from the point of view of information theory, there are several theoretical frameworks that appear to be related to this effect:
\begin{itemize}
    \item Structural Causal Models \cite{peters_elements_2017} consist of
      families of random variables linked by certain relationships that
      (implicitly) involve a notion of time: consider random variables $X_1,
      \ldots, X_n$ such that for each $i\geq 1$, $X_{i + 1}$ is a written as $
      f_i(X_1,\ldots, X_{i}, Z_i) $, where the $Z_i$'s are jointly independent.
      While the presence of such a decomposition is not special to the order in
      which the variables are labeled, an order may be singled out in certain
      cases if we put constraints on the structure of $f_i$ and $Z_i$ (e.g.
      that $f_i, Z_i$ are `simple' in some sense): if we e.g. reverse the order
      of the variables $(\tilde X_1,\ldots,\tilde X_n)=(X_n,\ldots,X_1)$, it
      may not be possible to write $\tilde X_{i+1} = \tilde f_i (\tilde
      X_1,\ldots, \tilde X_i, \tilde Z_i)$, with $ \tilde f_i, \tilde Z_i $
      being as `simple' as $f_i, Z_i$. This may have an impact on learnability
      \cite{scholkopf_towards_2021}, akin to that of an AoT. 
    \item Computationally-constrained views on information theory, in
      particular the recent framework of \emph{$\mathcal V$-information}
      \cite{xu_theory_2020}, allow one to take into account the computational
      challenges associated with the information extraction; through the lens
      of the $\mathcal V$-information, we see the emergence of symmetry
      breaking, an example of which is the Arrow-of-Time effect.
\end{itemize}

\section{\label{sec:empirical-results-on-nat-lang}Empirical Results on Natural Language}

In this section, we explore the existence of an AoT for LLMs in natural
language datasets. We first reveal the presence of an AoT in a basic setup
(GPT2 models on English and French with context window of length 256, see
Section \ref{subsec:emp-res-eng-fr} below). We then decline our explorations
over more than 50 model modalities (GPT/GRU/LSTM architectures, 6 GPT sizes,
context window lengths of 16/32/64/128/256/512 and 8 languages), and rule out
possible tokenization artifacts. We observe a consistent FW AoT in these
setups, including a number of takeaways concerning its magnitude.

\subsection{\label{subsec:emp-setup}Setup}

For the identification of an AoT in a dataset, we make sure that both the FW
and BW models are trained with the exact same specifications: the only reason
for a difference between the models' performances is hence the token prediction
order. In all experiments, the models are trained from scratch, using He
initialization \cite{he_delving_2015}.

\subsubsection{Dataset and Tokenization}

We conduct our natural language experiments on the CC-100 dataset
\cite{wenzek_ccnet_2019, conneau_unsupervised_2020}, which provides large
monolingual text datasets for various of languages and is reasonably
homogeneous across languages. This dataset is made of Commoncrawl snapshots, filtered 
for quality by comparing the data with a Wikipedia-trained model \cite{wenzek_ccnet_2019} (use the 
\href{https://huggingface.co/datasets/cc100/viewer}{Huggingface viewer} to explore the dataset). For
each language, we train from scratch a BPE tokenizer
\cite{sennrich_neural_2016}, with a vocabulary size of 50257, the same as GPT2
\cite{radford_language_2019}, including the beginning of sentence
$\langle$BOS$\rangle$ token.

To train on length-$n$ data batches, we split the dataset into `sentences' of
$n-1$ tokens, with a stride of $\frac{n}{2}$, ensuring that each token can be
seen at least once with reasonable context. For the FW model, we add the
$\langle$BOS$\rangle$ token at the start of the sequence, while for the BW
model, we add the $\langle$BOS$\rangle$ token at the end, and flip the
token order before feeding it to the model. We withhold $\sim 250k$ sentences from
the dataset for validation.

\subsubsection{Models, Hyperparameters and Training}

While some experiments involve other autoregressive models (GRU, LSTM), for
most training jobs we use the decoder-only Transformer (GPT)
\cite{radford_improving_2018}; our implementation (with code in the
supplementary material) is derived from minGPT
\cite{andrej_karpathymingpt_2024}. All GPT experiments use learned positional
embeddings and a dropout rate of $0.1$. Other hyperparameters may depend on the
experiment, see the Appendix.

For all models, we use the AdamW optimizer \cite{loshchilov_decoupled_2019}
with base learning rate of $10^{-4}$ and a learning rate schedule with a
warmup, followed by cosine annealing with warm restarts
\cite{loshchilov_sgdr_2017}. These
hyperparameters are mostly kept constant across different experiments, although
the period of the warm restarts might be tweaked to synchronize the end of
training with the end of a cycle, see Appendix \ref{app:natural-language-details} for details.

\subsection{\label{subsec:emp-res}Results}

In this section, we present results of various experiments confirming the
presence of a FW AoT in English and French datasets, and provide compelling
evidence for its universal existence in natural languages by considering six
other languages (five distinct families in total).

\subsubsection{\label{subsec:emp-res-eng-fr}Arrow of Time in English and French}

We begin in this section by analyzing the difference in FW vs BW training
dynamics for a Transformer of size GPT2-Medium ($\sim405M$ parameters,
context window length of $256$) on the CC-100 datasets for English and French. We
train the FW and BW models for the equivalent of 1 epoch of the French dataset
($\sim27B$ tokens), avoiding memorization.

\begin{figure}[h!]
\centering \includesvg[width=\columnwidth]{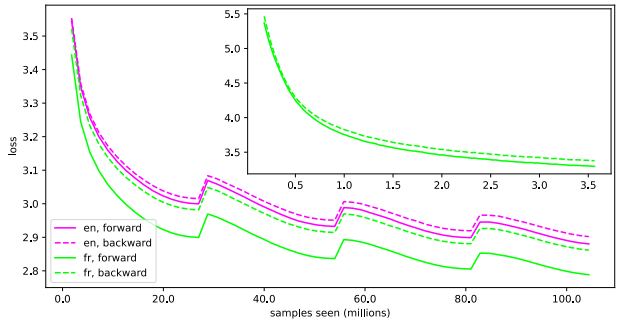}
\caption{English vs French validation losses (French training losses in the zoom-in, early loss values cropped for readability).}
\label{fig:envsfrench}
\end{figure}

As is seen in the zoom-in of Fig 1., after an initial short transition period,
the BW model loss separates from its FW counterpart and settles slightly above
it, and then follows an almost parallel trajectory. This consistent difference
throughout training (even persisting through warm restarts) points to the
existence of an AoT both in English and in French: at the end of training, we see the following losses for English: FW: $2.88$, BW: $2.902$ a difference of $+0.76\%$; and for French: FW: $2.788$, BW $2.862$, a difference of $+2.65\%$.
Interestingly, the magnitude of this effect is
different for English and French.

As will be discussed in the
next subsections, the findings are quite universal: they can be consistently
expanded to various settings, across models, languages, and context lengths.

\subsubsection{Context Window Size}\label{subsec:context-window-size}

In this section, we examine the influence of long-range correlations on the
AoT, by studying its relationship with the context length. Intuitively, for a
very small context length, we should see virtually no AoT; with very few
tokens, models approach the optimal solutions similarly, as they have
fewer degrees of freedom. For instance, in the extreme case of a context of
length 2, models are only tasked with learning a two-variable function
$\mathcal{V}^{2}\to\left[0,1\right]$, i.e. to learn the frequencies of 2-grams,
which should be (equally) easy in both directions. It is likely that an AoT
emerges for larger context lengths (and for reasonably large models).

\begin{figure}[h!]
\centering \includesvg[width=1.0\linewidth]{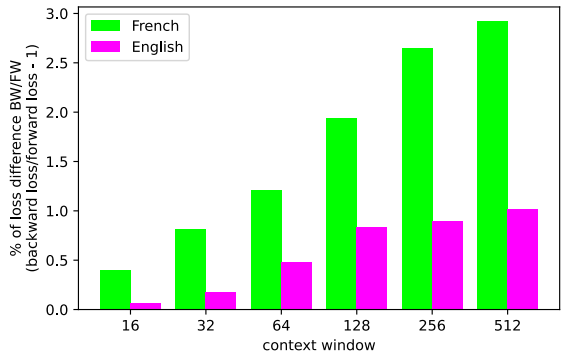}
\caption{BW/FW losses percentage difference for different context lengths}
\label{fig:context-lengths}
\end{figure}

We test
the dependence on the context window by training the same GPT-Medium model, but
with context lengths spanning from $16$ to $512$ tokens, both on English and French.
As can be seen in Fig. \ref{fig:context-lengths}, the magnitude of the AoT in both
English and French increases with the context size, suggesting the importance
of long-range dependences.

\subsubsection{Model Size}

In this section, we investigate the effect of model size for GPT models (other models are discussed in the next subsection). As in Section \ref{subsec:context-window-size} above,
it is natural to expect small models to struggle to exhibit an AoT that would
depend on sophisticated, long-range dependences. To test this, we train GPT
models of different sizes, from $5M$ to $405M$ parameters, all with a context length
of $256$. Interestingly the AoT is much smaller at very small model sizes,
reinforcing the idea that long-range dependences are key; as the model size
keeps growing beyond that, the difference tends to grow. Note also that larger BW models typically outperform smaller FW ones.

\begin{table}[h!]
\caption{Final FW losses and relative BW differences.}
\centering \resizebox{\columnwidth}{!}{%
\begin{tabular}{|l|l|l|l|l|l|l|l|}
\hline
 & Nano  & Micro  & Mini  & Small  & GPT1  & Med \tabularnewline
Size  & 4.9M  & 13.7M  & 22.0M  & 55.6M  & 162M  & 405M \tabularnewline
\hline
Fr-FW  & 4.525 & 3.964 & 3.683 & 3.293 & 2.979 & 2.788 \tabularnewline
Fr-BW & +0.15\% & +0.63\% & +1.49\% & +1.64\% & +2.07\% & +2.65\% \tabularnewline

\hline
En-FW  & 4.599 & 4.064 & 3.799 & 3.416 & 3.081 & 2.880 \tabularnewline
En-BW  & -0.33\% & +0.1\% & +0.11\% & +0.26\% & +0.49\% & +0.76\% \tabularnewline
\hline
\end{tabular}}
\end{table}
\subsubsection{Other Models}

While most results in this paper are focused on GPT models (the current state
of the art for language modeling), the question of the AoT can naturally be
asked for other autoregressive models. We investigate this for GRUs and LSTMs (three
sizes each), again with a context length of 256.

Once more, for sufficiently
large models, we observe a consistent AoT throughout modalities, confirming that the
observed AoT goes beyond Transformer models; rather, it appears to be intrinsic to the dataset. It is interesting for instance that for the English
dataset, the smaller BW model performs slightly better than the FW one. This
however convincingly disappears for larger context sizes and models.

\begin{table}[h!]
\caption{Final FW GRU/LSTM losses and relative BW differences.}
\resizebox{1.02\columnwidth}{!}{%
\begin{tabular}{|l|l|l|l|l|l|l|}
\hline
 & GRU S  & GRU M  & GRU L  & LSTM S  & LSTM M  & LSTM L \tabularnewline
Size  & 26.1M  & 56.2M & 94.9M  & 26.3M  & 57.8M  & 101M \tabularnewline
\hline
Fr-FW & 3.905  & 3.692 & 3.363  & 3.901 & 3.566 & 3.314 \tabularnewline
Fr-BW & +0.26\%  & +0.3\%  & +0.62\%  & +0.1\%  & +0.45\%  & +0.66\% \tabularnewline
\hline
En-FW & 4.030 & 3.712 & 3.483 & 4.015 & 3.653 & 3.418 \tabularnewline
En-BW & -0.07\%  & +0.22\%  & +0.34\%  & -0.27\%  & +0.11\%  & +0.15\% \tabularnewline
\hline
\end{tabular}}
\end{table}

\subsubsection{Other Languages}
\begin{figure*}[ht]
    \centering
    \vspace{-.2cm}
    \includegraphics[width=0.78\textwidth]{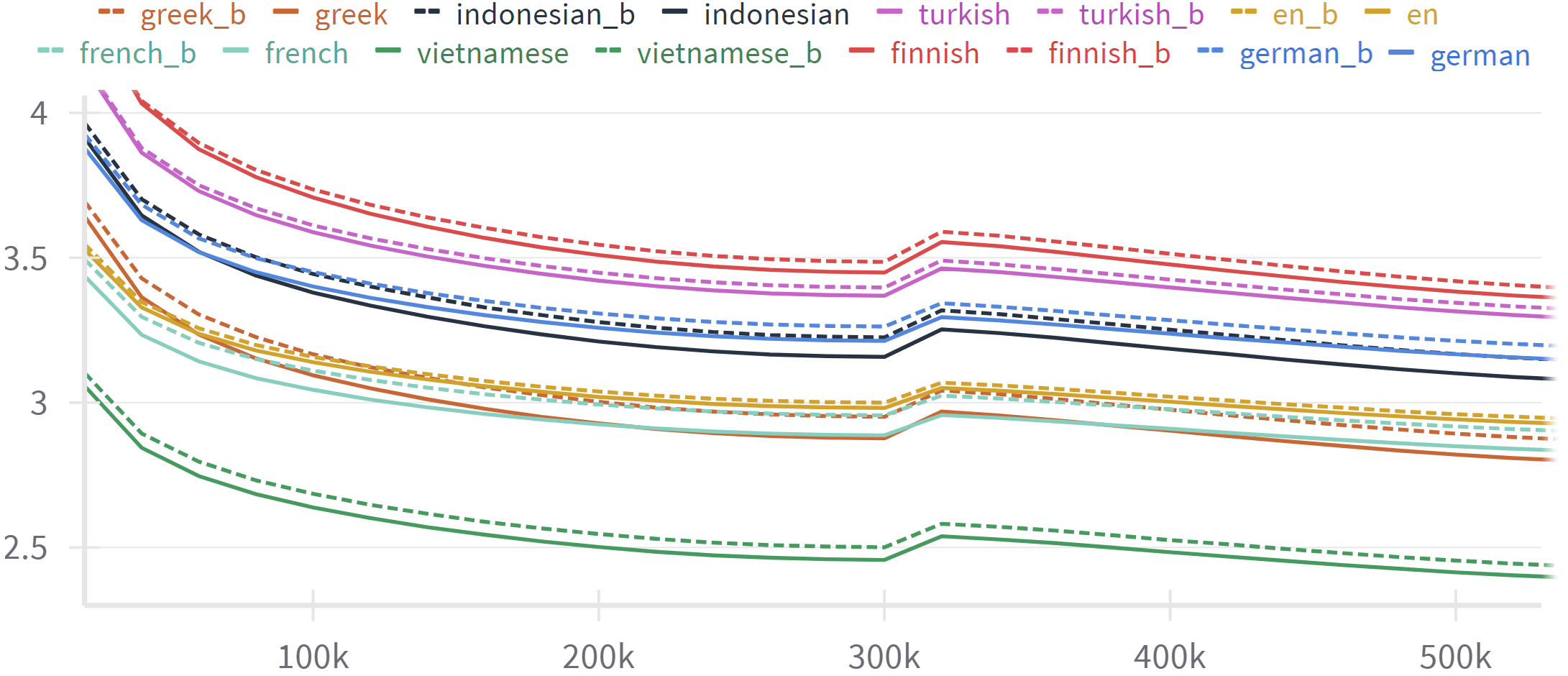}
    \vspace{-.1cm}
    \caption{Validation loss curves for FW and BW models during training. Consistently, the BW loss is higher than its FW counterpart. This persists through the warm restart of the learning rate, which causes a bump in the loss.}
    \vspace{-.3cm}
    \label{fig:all_languages}
\end{figure*}

The above experiments confirm the existence of an AoT
for English and French across various modalities. An exciting question
that naturally arises is whether this might be
a universal property of natural languages. To begin to explore this
question, we train models of two sizes (GPT2-Medium and GPT2-XL) on six
more languages.
\begin{table}[h!]
\caption{Final losses for different languages, Medium/XL models. Format:
{[}Final FW loss{]}/{[}BW relative difference{]}.}
\resizebox{\columnwidth}{!}{%
\begin{tabular}{|l|l|l|l|l|}
\hline
 & German  & Turkish  & Finnish  & Vietnamese \tabularnewline
\hline
Med  & 3.148/+1.46\%  & 3.292/+0.94\%  & 3.359/+1.07\%  & 2.396/+1.67\% \tabularnewline
\hline
XL  & 2.892/+1.59\%  & 3.084/+1.17\%  & 2.975/+1.85\%  & 2.099/+2.81\% \tabularnewline
\hline
\end{tabular}} \resizebox{\columnwidth}{!}{%
\begin{tabular}{|l|l|l|l|l|}
\hline
 & Greek  & Indonesian  & French  & English \tabularnewline
\hline
Med  & 2.794/+2.4\%  & 3.079/+2.18\%  & 2.834/+2.4\%  & 2.926/+0.61\% \tabularnewline
\hline
XL  & 2.494/+3.05\%  &  2.741/+3.17\%  & 2.586/+2.51\%  & 2.683/+0.63\% \tabularnewline
\hline
\end{tabular}}
\label{tab:other-languages}
\end{table}

From Table \ref{tab:other-languages}, we can see that in all the cases we
tested, a FW AoT emerges, although its magnitude appears to vary from language
to language, suggesting some universality of this phenomenon across human
languages. Fig. \ref{fig:all_languages} showcases the stability of this AoT
during training, across languages. Three more languages (Tagalog, Hebrew and
Arab) were tested at the suggestion of reviewers, confirming the universality
of the AoT in human languages (see Appendix \ref{app:other-languages}).

\subsubsection{Possible Artifacts}\label{subsec:tokenization-artifacts}

While the training procedures are perfectly symmetric with respect to the two
directions, it is important to rule out any other possible sources of
asymmetry. One possible source could in principle be the tokenization; indeed,
the BPE tokenizer is trained in the FW direction. To rule out this possibility,
we inverted (at character level) two datasets (Greek and French), and re-trained a BPE
tokenizer on the result. We trained a GPT2-Medium on it, and confirmed that the
direction of the BPE tokenization has no effect on the training dynamics: in
this case, the FW (respectively BW) model performs very closely to the
BW (respectively FW) model on the original tokenization, thereby showing
exactly the same AoT. See Appendix \ref{app:bpe-tok} for
details.

Additionally, one might ask about the variation in the final losses due to initialization. Although the agreement of all the different experiments show that this is negligible w.r.t. the AoT effect, we quantify this influence by repeating experiments for Greek, see app. \ref{app:greek-repeat}.

\subsection{\label{subsec:emp-takeways}Key Takeaways}
The above experiments suggest the universality of the phenomenon of
AoT across languages, models, and hyperparameters. More specifically:

\begin{itemize}
\item A very consistent AoT emerges for large enough models, trained for long enough, and with
  a large enough context window; in the other cases, the effects are less
  clear.
\item An important finding is that the magnitude of the AoT increases with the
    context length: this suggests the importance of long-range correlations;
    relatedly, the model's size can influence its ability to
    use the information of its whole context window.
\item While most of our training is done with GPT models, we observe the same
  type of results for GRUs and LSTMs, suggesting that AoTs are intrinsic to datasets.
\item An interesting phenomenon is that the magnitude of the AoT greatly depends
  on the language, even if its presence and direction are universal. Explaining this
  convincingly remains a fascinating challenge.
\end{itemize}
In Section \ref{sec:computability-and-irreversibility}, we introduce
a framework to reveal the emergence of the AoT in synthetic datasets
and propose mechanisms to explain how this can apply to natural languages.
In Section \ref{sec:discussion}, we discuss how these somehow surprising
results open the door to many possible investigations.

\section{\label{sec:computability-and-irreversibility}Computability and Irreversibility}

As discussed in \ref{subsec:cross-entropy-loss-and-perplexity} above, from an
information-theoretic point of view (abstracting away computability), there
should be no difference between FW/BW models. However, as shown in
\ref{subsec:emp-res}, we see a consistent AoT for various types of
architectures across multiple modalities, which increases with larger context
windows. As a result, any plausible explanation must explain why certain
probabilities are harder either to be (1) \emph{represented }or (2)
\emph{learned} with BW models than with FW ones. Naturally (1) is stronger than
(2): models cannot learn what they cannot represent (e.g. if there exists no set of model weights that solve the problem). In this section, we
provide simple mathematical models of data illustrating how both mechanisms can
arise and naturally contribute to the AoT. We start with a simple mathematical
model using prime number multiplications, illustrating how the computational
hardness of reversing certain information-preserving operations generates an
AoT. We then construct a more general class of data models based on binary
operations, allowing one to reveal an AoT based on sparsity and complexity
theory ideas.

\subsection{\label{subsec:number-factoring}Number Factoring and Arrow of Time}

Perhaps the most classical example of information-preserving, yet hard to
invert, computation is number factoring: given two large primes $p,q$ with
$p<q$, it is relatively easy to compute $n=pq$; while $n$ contains the same
information as $p$ and $q$, recovering them from their product is (believed to
be) very hard. This problem is the basis of much of asymmetric cryptography. In
this section, we study how FW/BW models perform when trained on a dataset based
on this idea; we study the theoretical entropy distribution when reading the
data FW and BW and compare this to the experimental values for FW/BW GPTs.

\subsubsection{Synthetic dataset}

For fixed $k\geq1$, consider the language of strings of the form $p\times
q\leftrightarrow\mathrm{rev}\left(pq\right)$, with $p<q$ primes, $p,q<10^{k}$
and $\mathrm{rev}\left(pq\right)$ being the product of $p$ times $q$, written
in reverse order (see \cite{shen_positional_2023, lee_teaching_2023}). The
numbers $p$ and $q$ are padded to be of $k$ digits exactly and the
$\mathrm{rev}\left(pq\right)$ is padded to be of $2k$ digits exactly (e.g. for
$k=4$: $0019\times0023\leftrightarrow73400000$). The symbols $\times$ and
$\leftrightarrow$ are written as multiple tokens (3 tokens for $\times$, and 7
tokens for $\leftrightarrow$), to facilitate the learning of non-trivial
operations by GPTs \cite{thomas_ahle_thomasahle_this_2023}. For a fixed $k$,
$\mathbb{P}$ is thus supported on $4k+10$ token sequences; in our experiments,
we set $k=5$ and take $10^{8}$ such random ordered pairs. Intuitively,
computing the right-hand side (RHS) of the symbol $\leftrightarrow$ given the
left-hand side (LHS) should be easy, while computing the LHS
from the RHS should be much harder (at least finding $q$; given $q$ and
$\mathrm{rev}\left(pq\right)$, finding $p$ should be easier).

\subsubsection{Nats of entropy}

In order to better understand the experimental results, we compute the
aggregate entropy (in nats) on $p$, $q$ and $\mathrm{rev}\left(pq\right)$ when
reading the strings $p\times q\leftrightarrow\mathrm{rev}\left(pq\right)$ FW
and BW (we do not compute the entropy on each token individually, and the
entropy on the symbols $\times$ and $\leftrightarrow$ is $0$). For instance,
for $k=5$, there are $\ln\left(\pi\left(10^{5}\right)\right)=9.17$ (with
$\pi\left(x\right)=\#{p:p\leq x}$) nats of entropy over the possible prime
numbers $<10^{5}$, which drops to $8.98$ nats of entropy on $p$ (because of the
ordering), and (averaging over $p$) $8.67$ nats of entropy on $q$; this results
in $17.64$ nats of entropy for the pair $\left(p,q\right)$, which is roughly
$2\times9.17-\ln\left(2\right)$ (we subtract the bit of information due to the
ordering); since $\mathrm{rev}\left(pq\right)$ is determined by $p$ and $q$,
its conditional entropy is naturally zero. Reading the string backward, the
$17.64$ nats of entropy are concentrated on $\mathrm{rev}\left(pq\right)$; the
rest is fully determined, and thus has zero entropy.

\subsubsection{\label{subsec:number-factoring-emp-res}Experimental Results}

Training a model with a GPT2-Medium on the $p \times q$ dataset yields the log-perplexities recorded in Table \ref{tab:primes}.
The FW model is able to reach the information-theoretical limits
on $p$ and $q$; the conditional cross-entropy loss on
$\mathrm{rev}\left(pq\right)$ is low but non-zero, indicating that the model
(imperfectly) learns to multiply the prime numbers. In contrast, the results for the BW model show a far-from-optimal perplexity on $\mathrm{rev}\left(pq\right)$, which points to the difficulty for the model to recognize the products of two primes; knowing $\mathrm{rev}\left(pq\right)$, almost no information on the prime factor $q$ is extracted: only $8.98-8.41=0.57$ nats, i.e. less than one bit. The `division' is much more learnable, with all but $0.02$ nats of information learned. All in all, the total FW log-perplexity is $22.2$ nats, while the BW one reaches $30.2$ nats.

\begin{table}[h!]
\centering
\caption{Final perplexities for the prime numbers dataset}
\label{tab:primes}
\begin{tabular}{|l|l|l|l|}
\hline
 & $p$  & $q$  & $rev(pq)$ \tabularnewline
\hline
FW  & 8.98  & 8.67  & 4.55 \tabularnewline
\hline
BW  & 0.02  & 8.41  & 21.56 \tabularnewline
\hline
\end{tabular}
\end{table}

\subsubsection{\label{subsec:number-factoring-discussion}Discussion}

The above setup shows a significant AoT for the $p\times
q\leftrightarrow\mathrm{rev}\left(pq\right)$ dataset. This discrepancy can be
largely attributed to the asymmetry between the difficulty of factoring versus
multiplication: compared to the information-theoretical limit, about $4.55$
nats are lost for the multiplication, while $8.43$ nats are lost for the
factorization. We also see that the different structures of the LHS and RHS
(which have the same information-theoretic content as they determine each
other) also present a significant difference w.r.t. the models' abilities:
while the FW model reaches essentially optimal perplexity for the LHS (i.e. it
recognizes primes $<10^{k}$), the BW model is very far from optimal on the RHS
(i.e. to recognize products of primes pairs $p,q<10^{k}$ proves to be much more
difficult). While part of the asymmetry is attributable to the models'
specifics, as long as the dataset size is kept high enough that all pairs
$\left(p,q\right)$ cannot be memorized, a significant AoT can be expected: as
the model size (and training time) grows, multiplication will eventually be
learned (long before the dataset can be memorized,
\cite{shen_positional_2023}), while extracting substantial information from
$\mathrm{rev}\left(pq\right)$ about $q$ should remain very hard. The above data
model displays an AoT of types (1) and (2) (see
\ref{sec:computability-and-irreversibility} above): certain features turn out
to be harder to learn for the BW models, while others simply are likely not
even representable by LLMs (of reasonable sizes), as the alternative would
yield an efficient factorization algorithm.

\begin{rem}
Note also that the above also illustrates the importance of long-range
dependences for the AoT: if the context length is kept e.g. significantly
below $k$, the FW model will have trouble saying anything about
$\mathrm{rev}\left(pq\right)$, as $p$ will already be forgotten when reaching
the RHS, shrinking its advantage over the BW model.
\end{rem}

\begin{rem}
In the above setting, we are rooted in the computational difficulty of the
inversion of a bijective function; note that the core of the argument is the
\emph{computational difficulty}, rather than bijectivity. Abstracting
computability issues, there is still no difference between FW and BW
perplexities for optimal predictors, even if a mapping is not injective, or if it
is not well-defined as mapping (e.g. if some random noise is added to it). The
invertibility merely helps us get a simple computation of the theoretical
entropies, and to pinpoint where each model performs suboptimally; it is
however not directly related to the presence of an AoT.
\end{rem}

\subsection{\label{subsec:binary-operations}Binary Operations}

The model of Section \ref{subsec:number-factoring} shows how an AoT
\emph{can appear} in a dataset: in that example, based on computational
complexity ideas, we could handcraft a synthetic dataset that is both
practically and theoretically harder for a BW model than a FW one. This still
leaves the question of \emph{why} an AoT would arise in a dataset such as
natural language, as in Section \ref{sec:empirical-results-on-nat-lang}, and
why in one direction rather than another, i.e. why FW models would consistently
outperform BW ones. In this section, we introduce synthetic datasets based on
operations on the space of $m$-bit registers identified with $\mathbb{F}_{2}^{m}$
($\mathbb{F}_{2}$ denotes the field of integers mod $2$). We focus on languages
based on $\mathbb{F}_{2}$-linear circuits and relate their learnability to
their sparsity, using this to explain a difference between
FW and BW learnability; we then provide a setup motivating the specific FW
direction of the AoT in natural languages; we then provide experiments
validating our framework; we conclude by discussing extensions to the nonlinear
case.

\subsubsection{\label{subsec:linear-sparse-circuits}Linear Sparse Circuit Dynamics}

Consider the class of measures $\mathbb{P}_{n}$ on a \emph{linear language}
formed by sequences of $n=2m+1$ tokens, of the form $x\leftrightarrow y$ where
$x,y$ are random uniform on $\mathbb{F}_{2}^{m}$, but related by a bijective
linear map; $\leftrightarrow$ is counted as a token. For each $\mathbb{P}_{n}$,
we can write $y=f_{\mathbb{P}_{n}}^{\rightarrow}\left(x\right)$ and
$x=f_{\mathbb{P}_{n}}^{\leftarrow}\left(y\right)$ for
$f^{\leftrightarrows}:\mathbb{F}_{2}^{m}\to\mathbb{F}_{2}^{m}$. We define the
\emph{sparsity} of a linear map $f:\mathbb{F}_{2}^{m}\to\mathbb{F}_{2}^{m}$ as
the proportion of zero entries of its matrix. We will (informally) say that a
matrix is \emph{sparse} if this proportion is relatively high, i.e. close to
$1$. Intuitively, the sparsity of $f^{\rightarrow}$ (resp. of $f^{\leftarrow}$)
is related to how easy it is to learn $\mathbb{P}_{n}$ (based on random data
samples) for a FW model (resp. for a BW model). For GPT predictors, this is
studied numerically in Section \ref{subsec:empirical-results-on-lin-lang}
below. For a linear language $\mathbb{P}_{n}$, an AoT will thus emerge if the
sparsities of $f^{\rightarrow}$ and $f^{\leftarrow}$ are significantly different.
It is common knowledge that the inverse of a sparse matrix is generally less
sparse (e.g. \cite{duff_direct_2017}, Section 15.6). This is the basis for
the following claim (verified numerically in Appendix
\ref{app:sparsity-levels}):

\begin{claim}
\label{claim:sparse-and-less-sparse} If $A$ is a sparse random $m\times m$
matrix in $\mathbb{F}_{2}$ conditioned to be invertible, the matrix
$A^{-1}$ has typically lower sparsity. Similarly, if we perturb a
invertible matrix $M$ by a random sparse matrix $A$, we have that
the corresponding perturbation of the inverse $\left(M+A\right)^{-1}-M^{-1}$
is typically less sparse than $A$.

This claim can be used to show that in natural settings, if we want
to condition e.g. on $f^{\rightarrow}$ being sparse, this will result
in an $f^{\leftarrow}$ that is comparatively less sparse, and vice
versa. In \ref{subsec:toy-comm-setup} below, we propose a communication
setup where the sparsity of $f^{\rightarrow}$ is naturally favored,
yielding a FW AoT as observed in the natural languages (see
\ref{subsec:emp-res} above).
\end{claim}

\subsubsection{\label{subsec:toy-comm-setup}A Simple Communication Setup}

In the previous section, we have shown that the emergence of an AoT is natural
in the sparse setting: if we e.g. condition $f^{\rightarrow}$ to be sparse,
this will yield an inverse $f^{\leftarrow}$ that is less sparse. To motivate
the importance of sparsity, and in particular of FW sparsity (for the presence
of a FW AoT), we give a simple communication setup.

Suppose Alice and Bob are (human) agents with FW predictors having learned a
common language $\mathbb{P}_{B}$, and Carol is an (alien) agent with a BW
predictor having learned $\mathbb{P}_{B}$ as well. Now suppose Alice wants to
teach Bob a new language $\mathbb{P}_{A}$ by sending him samples from
$\mathbb{P}_{A}$; how easy this is will typically depend on how far away
$\mathbb{P}_{A}$ is from $\mathbb{P}_{B}$, i.e. how sparse
$f_{BA}^{\rightarrow}=f_{A}^{\rightarrow}-f_{B}^{\rightarrow}$ is. Assume Alice
is only able to teach $\mathbb{P}_{A}$ to Bob if $f_{BA}^{\rightarrow}$ is
sparse enough (note that Alice needs to learn $\mathbb{P}_{A}$ herself, it is
  reasonable to assume that she will only be able to do so if
$f_{BA}^{\rightarrow}$ is sparse enough). Conditioning on Alice being able to
teach $\mathbb{P}_{A}$ will hence yield (with high probability) a FW AoT:
following Claim \ref{claim:sparse-and-less-sparse} above,
$f_{BA}^{\leftarrow}=f_{A}^{\leftarrow}-f_{B}^{\leftarrow}$ will be typically
less sparse than $f_{BA}^{\rightarrow}$ and $\mathbb{P}_{A}$ will be harder for
Carol to learn than for Bob. This will ultimately impact the language
structure: if e.g. $f_{BA}^{\leftarrow}$ was often sparser than
$f_{BA}^{\rightarrow}$, it would be profitable to `restructure' the language,
expressing $y\leftrightarrow x$ rather than $x\leftrightarrow y$. This suggests
that selection pressure may cause languages to evolve to take a form
where $f_{BA}^{\rightarrow}$ is often sparser than $f_{BA}^{\leftarrow}$ ,
yielding a consistent FW AoT.

\subsubsection{\label{subsec:empirical-results-on-lin-lang}Experimental Results}

\begin{figure}[h!]
\centering \includesvg[width=0.9\columnwidth]{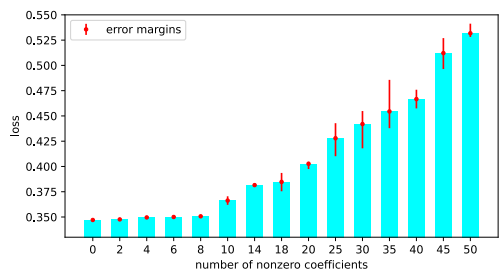}
\caption{Models loss at the end of training vs $f^{\rightarrow}$ sparsity.}
\label{fig:sparsities}
\end{figure}

In this section, we present experimental results supporting the claims of the
previous sections. We first consider a dataset made of linear languages with
$x,y\in\mathbb{F}_{2}^{25}$, for different sparsities of $f^{\rightarrow}$. We
train a GPT1 model on these datasets (see Appendix \ref{app:lin-langs}) and
plot the final losses in Fig. \ref{fig:sparsities}, confirming that sparser
matrices are easier to learn.

In the second experiment, we consider a model's ability to learn a sparse update,
given a learned prior: we first train FW/BW models on a linear
language with a $m=20$ sparse FW matrix, until it is learned perfectly, then
generate a new linear language by a sparse perturbation of the matrix. Table \ref{tab:modified-els} shows the losses of both models after 400 gradient descent steps.
Again, we see that the FW model adapts better to the sparse
modifications (see also Appendix \ref{subsec:sparse-updates}).

\begin{table}[h!]
\centering
\centering
\caption{GPT losses for various perturbations of the learned prior.}\label{tab:modified-els}
\resizebox{\columnwidth}{!}{
\begin{tabular}{|l|l|l|l|}
\hline
 & 2-bit flips  & 4-bit flips & 6-bit flips \tabularnewline
\hline
FW  & $0.347\pm 0.009$  & $0.354\pm 0.007$ & $0.367\pm 0.014$ \tabularnewline
\hline
BW  & $0.353\pm 0.008$  & $0.371\pm 0.011$ & $0.387\pm 0.013$ \tabularnewline
\hline
\end{tabular}}

\end{table}

\subsubsection{Non-Linear Case and Extensions}

If we consider a more general model of languages compared to Section
\ref{subsec:linear-sparse-circuits}, based on arbitrary functions
$f:\mathbb{F}_{2}^{m}\to\mathbb{F}_{2}^{m}$, the notion of sparse map needs to
be adapted to consider sparse \emph{circuits}, made of relatively few logic
gates (AND/OR/XOR) in the linear circuits. We can expect that for a random
sparse circuit model $f^{\rightarrow}$, the inverse (or any pre-image
computation) will typically be much less sparse. This is suggested by the fact
that inverting circuits is expected to be computationally hard (roughly the
content of the $P\neq NP$ conjecture). We can hence expect an AoT for similar
reasons. Due to the computational hardness of inverting nonlinear circuits, a
difference with the linear case can be expected to arise: the nature of the AoT in this case can
also be of Type 1 (see Section \ref{sec:computability-and-irreversibility}); in
some cases, a BW model may simply be unable to represent what the FW model
learns, as in the example of Section \ref{subsec:number-factoring}.

\section{\label{sec:discussion}Discussion}

In this paper, we have investigated the abilities of auto-regressive LLMs,
which predict tokens sequentially: for a given measure (dataset), we compare
the abilities of two models (FW/BW). Theoretically, if both models learn to
represent the same measure, their average log-perplexities should coincide. 
We discover the existence of an \emph{Arrow of Time (AoT)} for
natural language datasets: across a wide variety of models and hyperparameters,
all else being equal, FW models exhibit a consistently \emph{lower
perplexity} than BW ones; this difference emerges as soon as the model is
large enough; its causes appear rooted in long-range correlations in the data,
as the effect magnitude increases with the context length. 
We propose a framework to explain this phenomenon, based on
complexity and sparsity ideas: we construct examples of synthetic datasets
based on operations that display an asymmetry in terms of computability 
(despite being information-theoretically reversible); 
finally, we propose a setup where a FW AoT like the one seen in natural language can spontaneously emerge. 
Our work suggests a number of possible future research directions:
\begin{itemize}
\item Are AoTs universal across all human languages?
\item Are there AoTs in other
types of languages, e.g. computer code, binaries, DNA code, or bitmap
files?
\item How to explain the variation in magnitude of the AoT across languages?
\item Are there AoT scaling laws with respect to model sizes?
\item Are natural language AoTs of Type 1 or 2 (in the sense of Section \ref{sec:computability-and-irreversibility})?
\item For very long training times, is there a difference between train and test AoTs?
\item Can AoTs, Causality, and $\mathcal V$-Information (see Section \ref{subsec:relation-to-previous-theoretical-work}) be understood under a common framework?
\item What about AoTs in continuous settings, e.g. for video?
\item Is there a link with other AoTs, e.g. in thermodynamics?
\item Is the presence of an AoT in data a sign of life or intelligent processing?
\item Can we generalize the idea of flipping the order of the tokens to
other permutations of the context window?
\item Are AoT and computational hardness deeply linked?
\end{itemize}

In conclusion, the concept of AoT appears to be related to subtle properties of
natural language data, revealed through their interplay with autoregressive
LLMs. This idea's applicability seems wide, and a promising new tool to reveal
the presence of deep structural features in data. Further study of its
theoretical origin could prove fruitful towards uncovering links between AoTs
and complexity theory.

\newpage

\subsection*{Acknowledgements}

The authors would like to thank St\'ephane d'Ascoli, Samy Bengio, Gloria
Capano, Diego Dorn, Franck Gabriel, Ron Maimon, Jacob Menick, Christos
Papadimitriou, Jo\~ao Penedones, Matthieu Wyart, Nicolas Zlatoff, as well as
the anonymous reviewers for interesting discussions and suggestions. Support
from the Blavatnik Family Foundation, the Latsis Foundation, the NCCR SwissMAP,
and from an EPFL FSB Seed Funding Grant are gratefully acknowledged.

\section*{Impact Statement}
This paper presents work with the goal of advancing the field of Machine
Learning, and our scientific understanding of language models. There are many
potential societal consequences of a greater understanding of this
discipline, and thus, indirectly, of our work, however, none of them feel
direct enough to be specifically highlighted here. 

\bibliography{Arrow-of-Time}
\bibliographystyle{icml2024}


\newpage
\appendix
\onecolumn
\section{Details on training on natural languages}
\label{app:natural-language-details}

In this appendix, we provide more details on the training of our models on
natural languages.

For any experiment, the precise hyperparameters used can be found in the
code repository found at
\href{https://github.com/frotaur/LLM-Arrows-of-Time}{github.com/frotaur/LLM-Arrows-of-Time}, under the folder `Training Parameters', in $.json$
format. Those files can also be used to reproduce any experiment using the
codebase, as explained in the README.md of the repository. See also \href{https://arrowsoftime.org/}{arrowsoftime.org} for other relevant links.

All experiments (save for the 512 context size) were run on a single A100 GPU,
adjusting the batch size to fit the available memory.

Concerning the shuffling of the dataset, we proceed for all the experiments as
follows: we begin by splitting the textual dataset into `sentences' of the
appropriate context size $n$, with a stride of $n/2$ (i.e., if we have a
context size of 4 and the text is \textit{ABCDEF}, this results in two
sentences, \textit{ABCD} and \textit{CDEF}). This is to ensure that all tokens
appear in the training data with at least some context. After that, we shuffle
the obtained sentences with a set seed. We withhold 250k sentences (1000
batches at batch size 250) for validation. The inversion of the tokens is made
at the level of each batch; in this way, when training, the FW/BW models see
the data in the same order, preventing the emergence of undesirable
differences.

\subsection{Model sizes}
Table \ref{tab:modelsizes} provides more detail on the model sizes used in the paper.
\begin{table}[h!]
\caption{Model sizes used throughout the experiments. $d_{embed}$: number of embedding dimensions. $n_{heads}$: number of attention heads. $n_{layers}$: number of transformer blocks (attention + MLP). \textit{parameters}: total number of parameters, including the last linear layer which projects on vocabulary size (commonly referred to as the `head'). }
\centering
\begin{tabular}{|c|l|l|l|l|l|l|l|}
\hline
\multicolumn{1}{|l|}{GPT2 model name $\rightarrow$} & Nano  & Micro & Mini  & Small & GPT1 & Medium & XL   \\ \hline
$d_{embed}$                                         & 48    & 128   & 192   & 380   & 768  & 1024   & 1600 \\ \hline
$n_{heads}$                                         & 3     & 4     & 6     & 10    & 12   & 16     & 25   \\ \hline
$n_{layers}$                                        & 3     & 4     & 6     & 10    & 12   & 24     & 48   \\ \hline
parameters                                          & 4.92M & 13.7M & 22.0M & 55.6M & 162M & 405M   & 1.6B \\ \hline
\end{tabular}
\label{tab:modelsizes}
\end{table}

In the MLP layer, all models have one hidden layer with $4*d_{embed}$ hidden dimensions, a.k.a. an MLP ratio of $4$.

\subsection{Different languages}

\subsection{Context Window size}

For the testing of the influence of the context window length, we use a
GPT2-Medium model, with context window lengths going from $16$ to $512$ tokens.
Because of the different context lengths, it will take a model with a small
context length many more gradient steps to see the same amount of data. For
this reason, we do not train all models up to the equivalent of 1 epoch of the
French dataset ($\sim 26.6B$ tokens), but rather train them for sufficiently
long so that the perplexity differences stabilize, and that their losses
converge. Due to the cosine annealing learning rate schedule, we stop the
training at the end of a cosine decay, to avoid the bump in the loss caused by
a warm restart (note however that this has little to no effect on the
perplexity differences).

In Table \ref{tab:context_train_time}, we record the number of steps (i.e.
minibatches) that were seen for each context length.

\begin{table}[h!]
  \caption{Number of batches seen during training for the different context lengths. Note that the recorded batch size is the `effective' one, that is, potentially obtained through aggregation of smaller batch sizes.}
\centering
\begin{tabular}{|l|l|l|l|l|l|l|}
\hline
Context length $\rightarrow$ & 16    & 32    & 64    & 128   & 256  & 512  \\ \hline
batch size                   & 500   & 500   & 500   & 400   & 180  & 156  \\ \hline
seen batches                 & 1.74M & 1.62M & 1.64M & 1.02M & 0.5M & 0.6M \\ \hline
\end{tabular}
\label{tab:context_train_time}
\end{table}

For this experiment, because of our memory limitations, we trimmed down the
English dataset, which was too big when working with context windows of lengths
$16$ and $32$. Note that the 256 context length experiment in this section is
thus slightly different from the one recorded in Table
\ref{tab:other-languages}, due to the different datasets (as well as the batch size
and the cosine annealing period).

\subsection{Other Models}

The GRU and LSTM implementations we used for these experiments are those
natively implemented in the \verb|pytorch.nn| module of the Pytorch Python
library. In each case, we choose the `input size' (i.e., the number of
embedding dimensions for each token) to be equal to the `hidden size' (i.e.,
the number of hidden dimensions in the RNN's hidden state).

\begin{table}[h!]
\centering
\caption{Parameters for different sizes of LSTM and GRU models.}
\begin{tabular}{|l|l|l|l|}
\hline
               & Small & Medium & Large \\ \hline
hidden$_{dim}$ & 256   & 512    & 768   \\ \hline
n$_{layers}$   & 1     & 3      & 5     \\ \hline
\end{tabular}
\end{table}

Although RNNs have technically no limit on the context length, for training
purposes, to allow for a backward pass, we feed them with batches of texts of
lengths $256$.

\subsection{Other Languages}
\label{app:other-languages}

\begin{figure}[h!]
    \centering
    \includegraphics[width=.8\columnwidth]{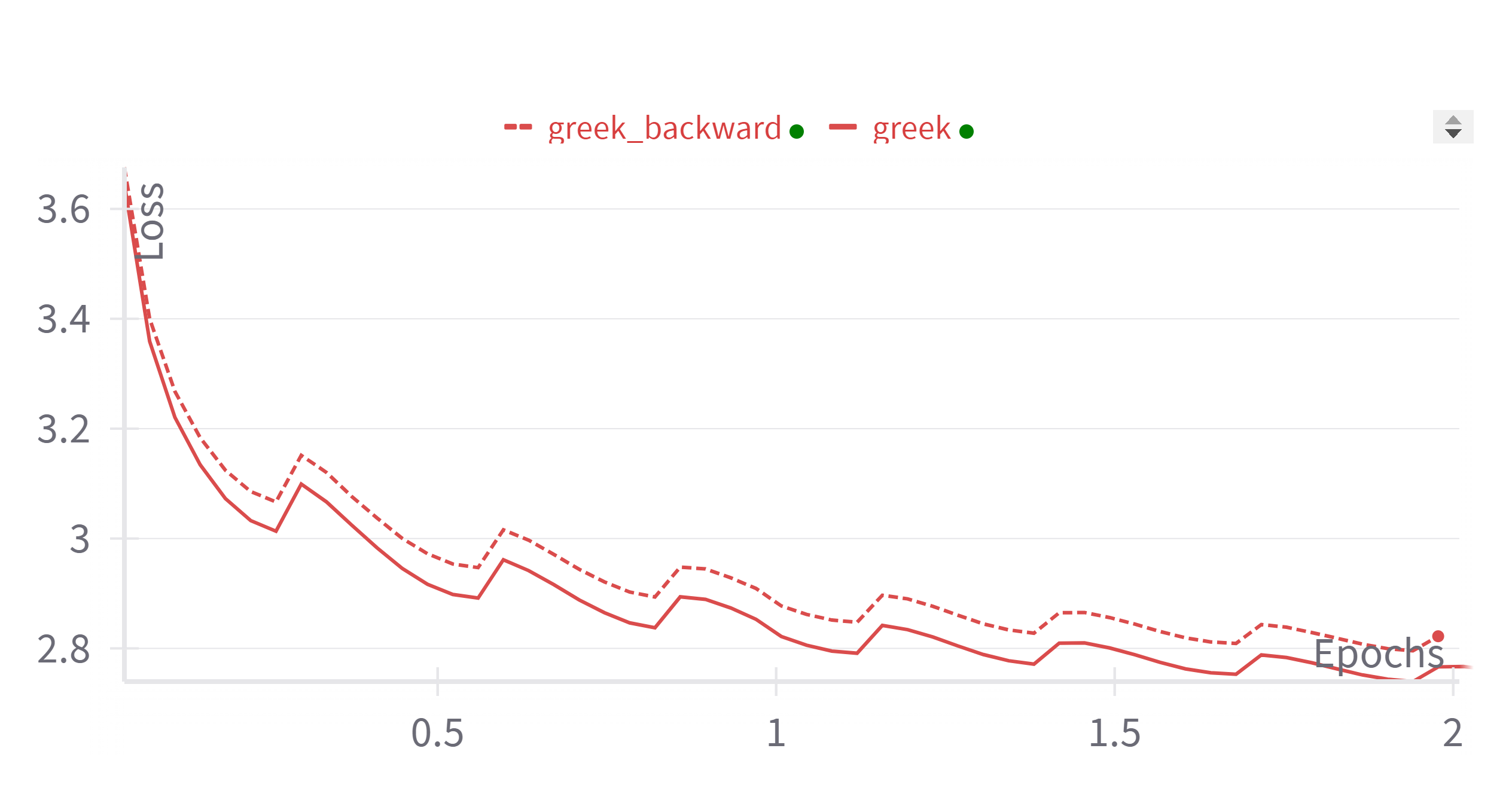}
    \caption{Validation loss for two epochs of training on the greek dataset, for forward and backward models.}
    \label{fig:greek-2-epochs}
\end{figure}

For the experiments on other languages, we decided to stop the training at the
equivalent of 1 epoch of Greek, which was one of the smallest datasets on which
we were training (along with Turkish). This choice was maintained for all
languages, and models never saw the same datapoint more than once (note that we also tried
training on Greek for two epochs, as shown in Fig. \ref{fig:greek-2-epochs};
this suggests that our results remain valid beyond one epoch).

For GPT2-Medium, a batch size of $90$ was used\footnote{Due to an oversight in
the code, the batches were not aggregated in groups of 2 as expected, but the
loss was still renormalized by dividing it by 2. This amounts to a very slight
change in learning dynamics, but does not affect any of the results. Similarly,
all graphs/reported results display the correct loss. In case one wants to
reproduce exactly the results of the paper, the loss should be divided by 2
before backpropagation.}, with one warm restart during training. For the XL
models, a batch size of $26$ was used, aggregated $6$ times for an effective
batch size of $156$, which didn't allow for a warm restart before the one epoch
of Greek.

We also tested (using GPT2-Medium, with an adjusted learning rate schedule, due
to the smaller size of the datasets) Hebrew, Arabic and Tagalog (see Table
\ref{tab:extra-languages} for final losses), at the suggestions of anonymous
reviewers. 

Hebrew and Arabic constituted an example of languages written right-to-left; a
priori, we would not expect this to affect the emergence of an AoT: after all,
tokens are still processed in the `spoken' order by the model, so the writing
direction does not affect training. Still, there could have been an influence
on the language itself from the writing direction, which cannot be detected
from our results.

Tagalog is an example of a language with `verb-initial word order', a
relatively rare class of languages, which was not included in our list. Here,
our expectations are again confirmed as an AoT appears also for this example.
This reinforces the idea that the AoT for natural language emerges from
long-range correlations. The specifics of the grammar and the order of words in
a sentence are therefore not that important.

\begin{table}[h!]
\caption{Final losses for extra languages. Arabic and Hebrew are at 1 epoch of training, and Tagalog at 7 epochs, due to the small size of the dataset. Format:
{[}Final FW loss{]}/{[}BW relative difference{]}.}
\resizebox{0.6\columnwidth}{!}{%
\begin{tabular}{|l|l|l|l|l|}
\hline
 & Tagalog  & Arabic  & Hebrew  \tabularnewline
\hline
Med & 2.368/+1.48\%  & 3.446/+1.91\%  & 3.288/+2.37\%   \tabularnewline
\hline
\end{tabular}}
\centering
\label{tab:extra-languages}
\end{table}
For completeness, we also attempted to run the training for Greek for 2 epochs, to see if memorization of the dataset may affect the Arrow of Time. In Fig. \ref{fig:greek-2-epochs}, we display the validation loss during training. Comparing the difference in performance at 1 and 2 epochs, it remains almost exactly the same. It would be interesting to test this with bigger models, and several epochs of training.

\subsection{BPE Tokenization}\label{app:bpe-tok}

For the tokenization, we use the Huggingface implementation of the BPE
Tokenizer (\href{https://huggingface.co/docs/tokenizers/api/tokenizer}{link}),
using the method \verb|Tokenizer.train_from_iterator|. The tokenizers are
trained on the same CC-100 dataset on which we train the model.

To exclude potential tokenization asymmetries (see section \ref{subsec:tokenization-artifacts}), we perform extra experiments in which we train the BPE tokenizer in reverse.  To do so, we reverse the language dataset at
the character level (not at the byte level, as this would make the output of
the model unreadable because of multi-byte characters in utf-8), then train the
BPE tokenizer on this new dataset. We then train a FW and a BW model on this
character-flipped dataset, tokenized with the new BPE tokenizer.

\begin{figure}[h!]
    \centering
    \includegraphics[width=0.8\linewidth]{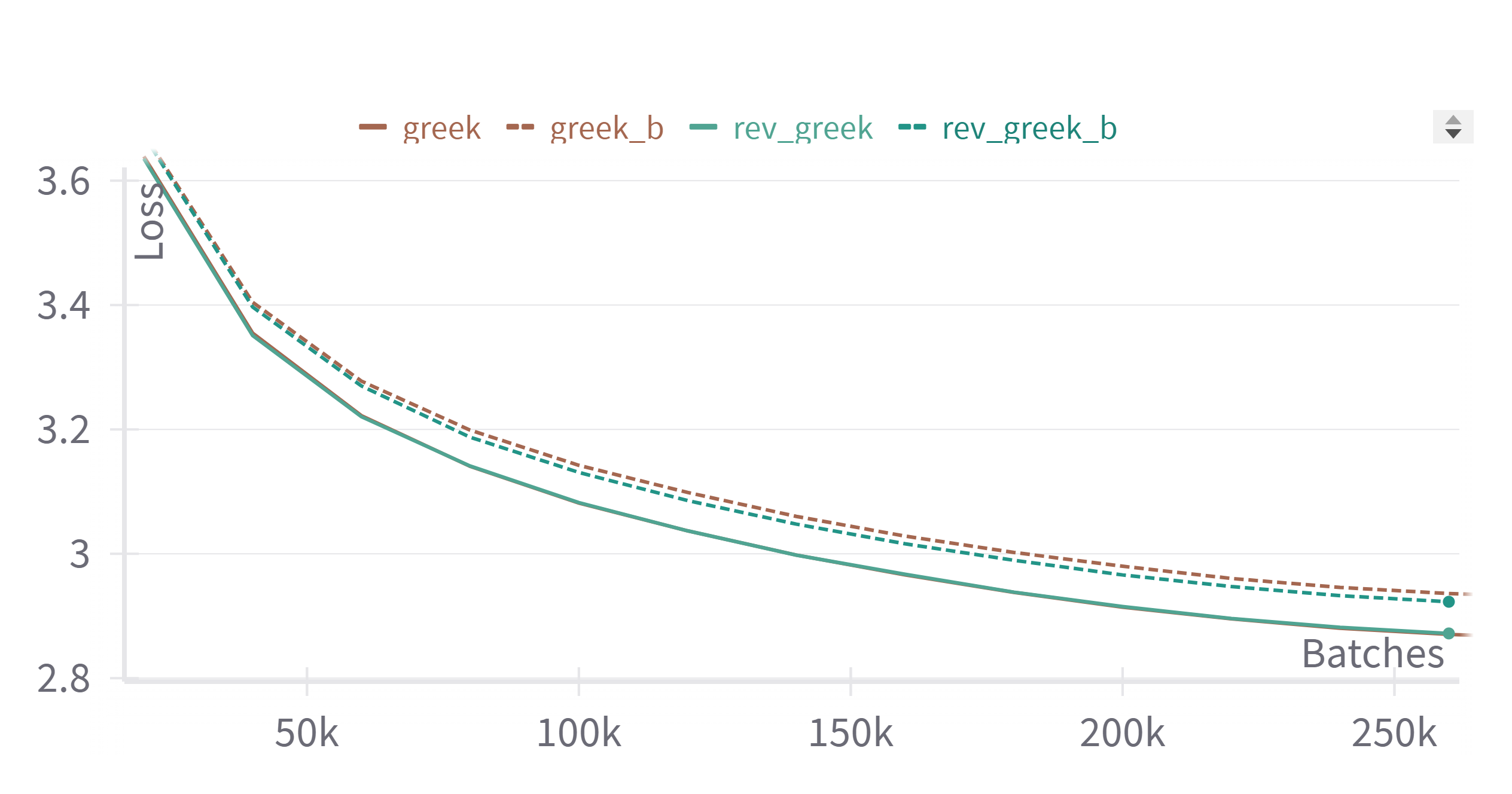}
    \caption{Validation curves for models training on the Greek dataset on forward and backward (label prefixed with `rev') BPE tokenizations. The Arrow of Time remains the same in the FW direction, despite the different tokenization schemes.}
    \label{fig:bpe-fw-bw}
\end{figure}

To make things clearer, we will call a model `backward' (BW) if it processes tokens in the opposite
order w.r.t. the natural reading direction (hence the `previous-token
predictor' on the `character-flipped dataset' corresponds to what we will call
the FW model). Thus, if the Arrow of Time is a property of the language and not
a tokenization artifact (as we expect), we expect the arrow of time to be in the same direction, independently of the tokenization scheme. Fig.
\ref{fig:bpe-fw-bw} confirms this; relying on the reverse BPE tokenization
introduces very slight differences in the losses. This difference is negligible compared to the AoT effect in both Greek and French,
so we can conclude that the AoT is not a tokenization artifact.

\begin{figure}[h!]
    \centering
    \includegraphics[width=0.8\linewidth]{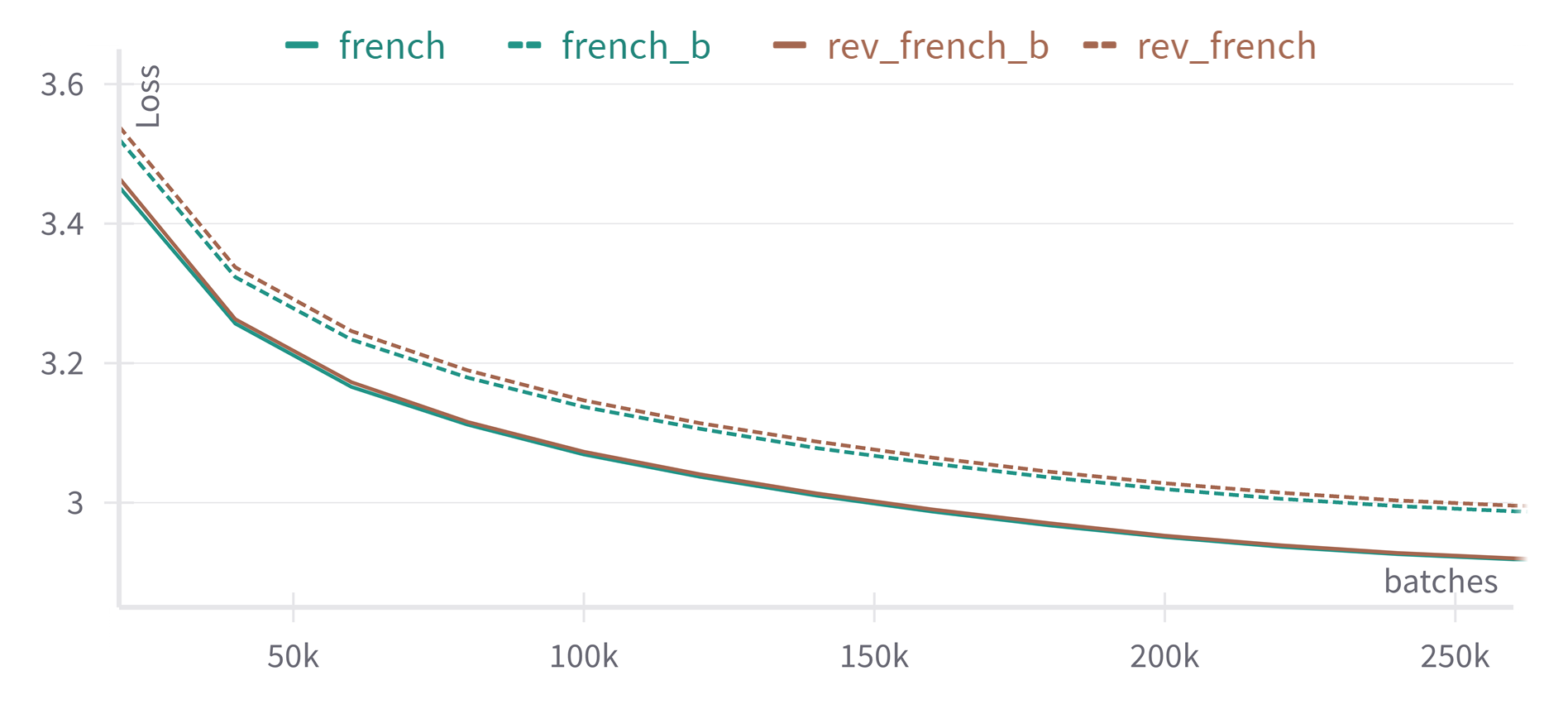}
    \caption{Validation curves for models training on the french dataset on forward and backward (label prefixed with `rev') BPE tokenizations. The Arrow of Time remains the same in the FW direction, despite the different tokenization schemes. It seems the reverse tokenization is slightly suboptimal, slightly degrading the losses of both FW and BW models.}
    \label{fig:bpe-fw-bw-fr}
\end{figure}

In figures \ref{fig:bpe-fw-bw} and \ref{fig:bpe-fw-bw-fr}, we display the loss curves during training for the french and greek models, trained using both the normal and reversed BPE tokenizers. The tokenizer switch has minimal impact on the losses, and most importantly, it does not affect the AoT direction.

\subsection{Initialization variance}\label{app:greek-repeat}

The AoT computed in our experiments were obtained with a single training run.
One might ask if such a difference remains significant compared to variations
in loss due to initialization. The consistency throughout experiments shows
that the AoT is significant, but in this section we set to verify the magnitude
of the initialization variance. Due to computational costs, it is not possible
to obtain error bars for all the experiments. Instead, we focus on Greek, for
which we re-run the GPT2-Medium training 4 times in total. Computing the error
bars, we obtain a loss (at one epoch) of $2.802\pm 0.005$ FW, and $2.871 \pm 0.003$ BW. This gives us an AoT magnitude of $2.46\% \pm 0.18 \%$, where we can see that the variations due
to initialization are negligible w.r.t. the magnitude of the AoT.

\section{Linear Language Toy Model}

\subsection{Matrix Inverse Sparsity}

Here, we substantiate the assertions of Claim
\textcolor{red}{\ref{claim:sparse-and-less-sparse}} by running numerical
experiments. To this end, we wish to look at $n\times n$ matrices in
$\mathbb{F}_2$ with a given number of non-zero elements (which we will call
$k$), and compute $k$ in the inverse. To generate invertible matrices with
extremely high sparsity (low $k$), we proceed as follows. We start with the
identity matrix $Id$, which is the only invertible matrix (up to permutations,
which do not affect sparsity) when $k=n$. To generate a matrix with
approximately $k$ non-zero elements, we flip $k-n$ elements of $Id$ at random.
This allows us to often obtain invertible matrices when $k$ is close to $n$,
which would not be the case if we simply selected the $k$ non-zero elements at
random. When $k$ becomes bigger than $n$, the initial presence of the identity
is quickly erased.

\begin{figure}[h!]
    \centering
    \includesvg[width=0.7\linewidth]{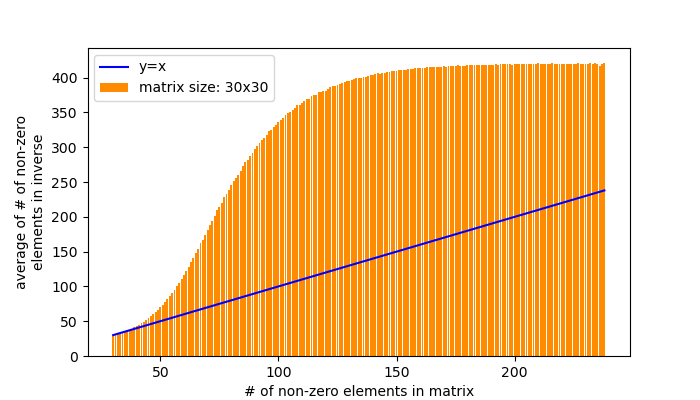}
    \caption{Plot displaying the connection between the sparsity of a matrix and its inverse. It is clear that on average, the inverse of a sparse matrix is less sparse. The number of non-zero elements for the inverse caps at $~450$.}
    \label{fig:sparsity-histo}
\end{figure}

We choose $n=30$, generate matrices with $k=[30,250]$, and record the average number of non-zero elements in the inverse, given $k$ fixed. Fig. \ref{fig:sparsity-histo} clearly confirms that the sparsity of the inverse is generally much lower.

\subsection{Linear Languages Experiments}
\label{app:lin-langs}

In this section, we give more details on the experiments of Section \ref{subsec:empirical-results-on-lin-lang}.

\subsubsection{Linear Language Dataset}
\label{sec:linear_language}

Given a matrix $M$ of size $n\times n$, the associated linear language dataset
will contain sentences of $2n+7$ tokens in the form $x\_\_\_\_\_\_\_y$, where
$x,y\in \mathbb{F}_2^n$, and the underscores are added as padding, providing
the model with more tokens if needed to perform more complex computations. The
vector $x$ is drawn at random, and $y$ is computed with $y=Mx$. We finish by
randomly flipping each bit with probability $p=0.01$ (which we call `adding'
perturbations), aimed at smoothing out the probabilities output by the model.
This is necessary for the `fine-tuning' experiments, as otherwise the models
become too confident in their predictions, and any small change $M$ results in
a huge change in the loss, leading to a catastrophic forgetting of the learned
prior.

\subsubsection{Sparsity Levels}
\label{app:sparsity-levels}

\begin{figure}[h!]
    \centering
    \includegraphics[width=0.7\linewidth]{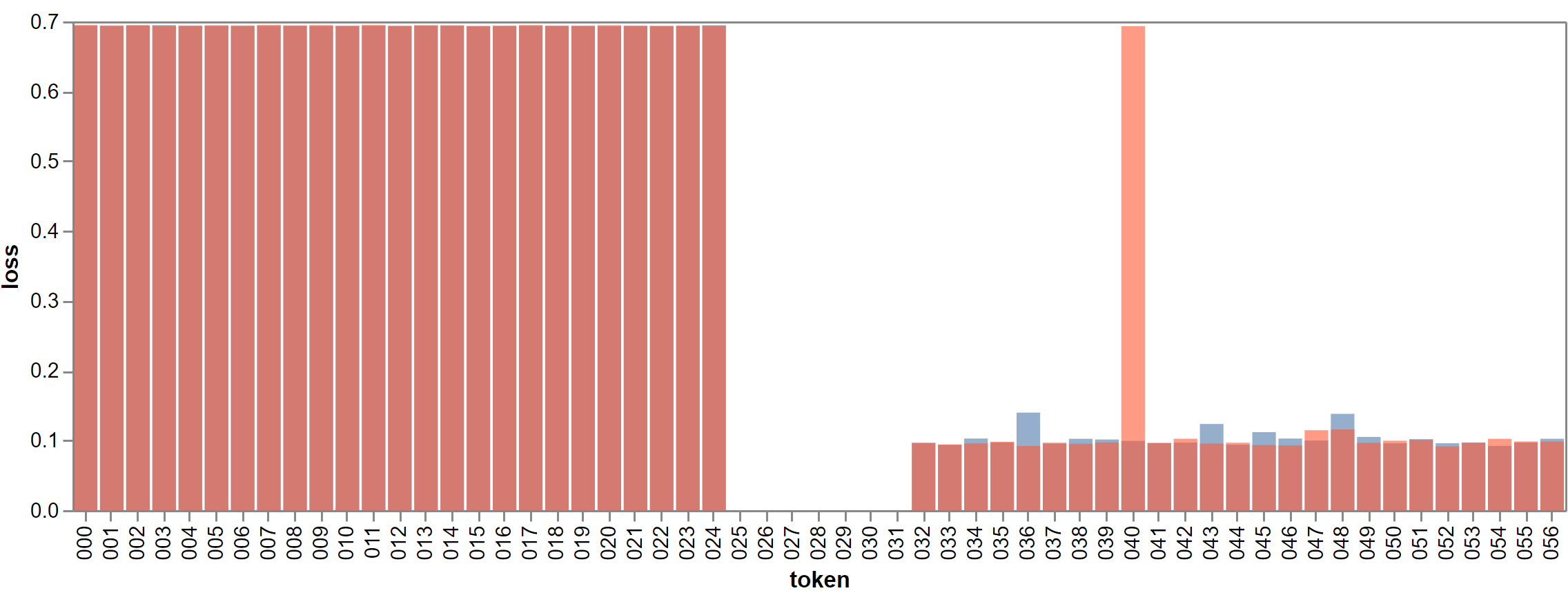}
    \caption{Average perplexities for each token in the linear language after $600k$ sentences, for two runs with $k=4$. One of the models is very close to the optimal solution, while the other is missing a single token. It usually takes a long time for the model to correct this, leading to higher variance in the final losses.}
    \label{fig:binary-learning}
\end{figure}

In the first experiment, we generate a linear language model in
$\mathbb{F}^{25}_2$, with $p=0$ perturbations, and matrices with a number
$25+k$ of non-zero elements, where $k \in
[0,2,4,8,10,14,18,20,25,30,35,40,45,50]$. We then train a transformer model of
size GPT1 (see Table \ref{tab:modelsizes}) on $600k$ sentences, with batch size
$200$. Note that the context size of the model matches exactly the number of
tokens in one sentence. Final losses are reported in \ref{fig:sparsities}. Note
that for lower sparsities, the trend is not obvious: this is due to the high
variance in the final learning rate, as the learning of only a few non-zero
elements is binary, depending on the initialization, the model either learns
the matrix perfectly quickly, or it usually struggles to find the last few
non-zero elements. Fig. \ref{fig:binary-learning} displays this behavior in the
case $k=4$. Note that the perturbations somewhat reduce this behavior, but
don't cancel it completely.

\begin{figure}[h!]
    \centering
    \includegraphics[width=0.8\linewidth]{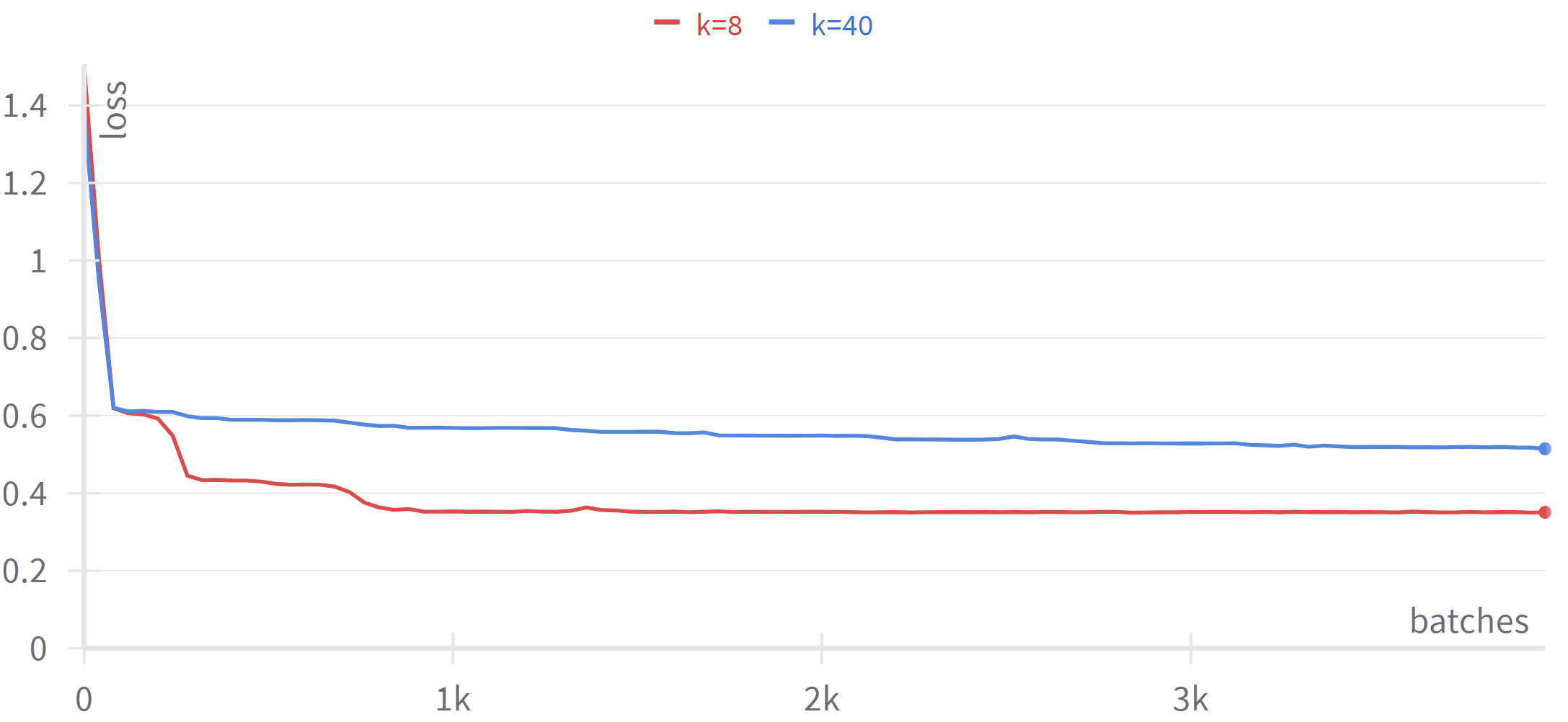}
    \caption{Loss during learning for $k=8$ and $k=40$ sparsities. The first plateau simply arises when the model learns to guess all coefficients randomly. Subsequently, the $k=8$ experiences plateaus each time it learns more non-zero coefficients. The learning of $k=40$ is much smoother, as discovering non-zero coefficients doesn't lead to perfect predictions right away.}
    \label{fig:40vs8}
\end{figure}

Fig. \ref{fig:40vs8} displays typical learning dynamics for this problem, for
$k=8$ (high sparsity) and $k=40$ (medium sparsity). We remind that in
principle, given a large enough model, and enough training steps, the model
should be able to find the optimal solution (hardness of type (2), see
Section \ref{sec:computability-and-irreversibility}).

\subsubsection{Sparse updates}\label{subsec:sparse-updates}

\begin{figure}[h!]
    \centering
    \includegraphics[width=0.85\linewidth]{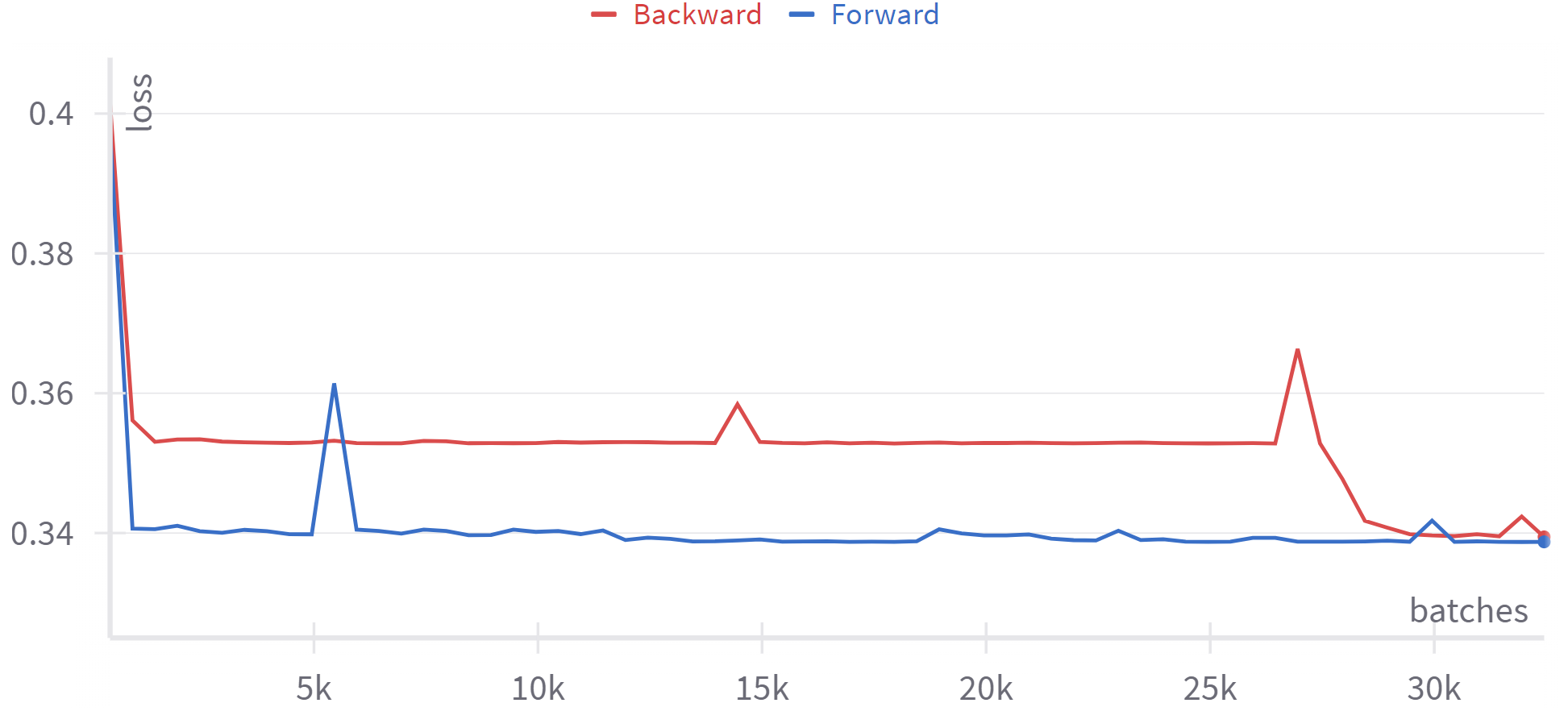}
    \caption{Training loss for the FW and BW models, trying to learn a linear language with a matrix $k=6$. While the FW model learns the quasi-optimal solution very quickly, the BW model remains stuck on a plateau for a long time. This in fact corresponds to a single element of the predicted vector which was missing, as in Fig. \ref{fig:binary-learning}.}
    \label{fig:fw-bw-perfect-learn}
\end{figure}

In the second experiment, we choose a linear language in $\mathbb{F}^{20}_2$. We
generate the dataset in the same way explained in Section
\ref{sec:linear_language}. We begin by training a FW model and a BW one on this
language, until both models learn it almost perfectly (batch size 200). For
this reason, we choose a very sparse matrix, with $k=6$ (in this specific
example, the inverse has $k=10$). As expected, this takes much longer for the
BW model, as displayed in Fig. \ref{fig:fw-bw-perfect-learn}.

\begin{figure}[h!]
    \centering
    \includegraphics[width=0.75\linewidth]{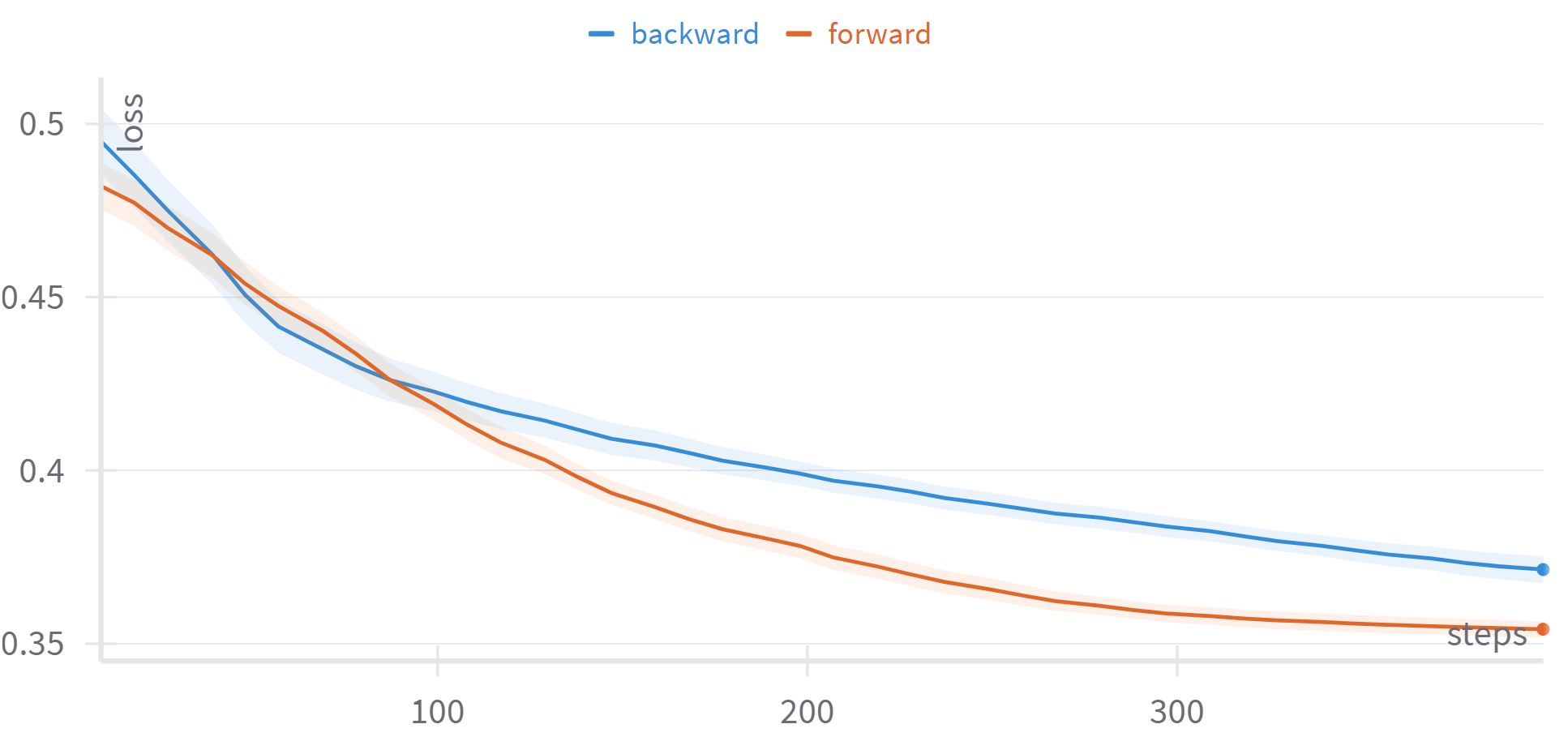}
    \caption{Averaged loss for forward and backward models, when trying to learn a sparse forward perturbation of the Linear language. In the first $\sim 100$ steps, the curves are similar as both models decrease their confidence in the new, perturbed tokens, setting them back to random chance. Subsequently, they begin learning the perturbation, where the forward model is clearly at an advantage.}
    \label{fig:training-sparse-pert}
\end{figure}

Once this `prior' is learned, we generate perturbation of the learned matrix by flipping $e$ entries of the matrix randomly, conditioned to the fact that it should remain invertible. We then train the models further on this new dataset, for a relatively small amount of steps (400 gradient steps). We also lower the learning rate to $8\times 10^{-6}$, with 10 steps of warmup, again to prevent catastrophic forgetting of the prior. The training dynamics are displayed in Fig. \ref{fig:training-sparse-pert}, in the case $e=4$.

We observe that the FW model adapts better than the BW one, and this is due to the fact mentioned in Claim \ref{claim:sparse-and-less-sparse}, namely that a sparse FW update will generically result in a less sparse BW update.
\end{document}